\documentclass[10pt,letterpaper]{article}
\usepackage{lmodern}
\usepackage[top=0.85in,left=2.75in,footskip=0.75in]{geometry}

\usepackage{amsmath,amssymb}

\usepackage{changepage}

\usepackage{textcomp,marvosym}

\usepackage{cite}

\usepackage{nameref,hyperref}
\usepackage[table]{xcolor}
\definecolor{bestgreen}{RGB}{198,239,206}
\definecolor{worstred}{RGB}{255,199,206}

\usepackage[nopatch=eqnum]{microtype}
\DisableLigatures[f]{encoding = *, family = * }

\usepackage[table]{xcolor}
\usepackage[most]{tcolorbox}
\definecolor{myblue}{HTML}{ddeef1}
\usepackage{array}
\usepackage{tabularx}
\usepackage{booktabs}
\newcolumntype{+}{!{\vrule width 2pt}}

\newlength\savedwidth

\raggedright
\usepackage[aboveskip=1pt,labelfont=bf,labelsep=period,justification=raggedright,singlelinecheck=off]{caption}

\makeatletter
\renewcommand{\@biblabel}[1]{\quad#1.}
\makeatother

\usepackage{algorithm}
\usepackage{algpseudocode}

\usepackage{lastpage,fancyhdr,graphicx}
\usepackage{epstopdf}
\fancyheadoffset[L]{2.25in}
\fancyfootoffset[L]{2.25in}
\begin{document}
\vspace*{0.2in}

\begin{flushleft}
{\Large
\textbf{MixSarc: A Bangla--English code-mixed corpus for implicit meaning identification}
}
\newline
\\
Kazi Samin Yasar Alam\textsuperscript{1},
Md Tanbir Chowdhury\textsuperscript{1},
Tamim Ahmed\textsuperscript{1},
Ajwad Abrar\textsuperscript{1*},
Md Rafid Haque\textsuperscript{2}
\\
\bigskip
\textbf{1} Department of Computer Science and Engineering, Islamic University of Technology, Dhaka, Bangladesh
\\
\textbf{2} Department of Computer Science, University of Illinois Chicago, Chicago, Illinois, United States of America
\\

\bigskip

%
%





* ajwadabrar@iut-dhaka.edu

\end{flushleft}


%
\section*{Abstract}
Bangla--English code-mixing is widespread across South Asian social media, yet resources for implicit meaning identification in this setting remain scarce. Existing sentiment and sarcasm models largely focus on monolingual English or high-resource languages and struggle with transliteration variation, cultural references, and intra-sentential language switching. To address this gap, we introduce MixSarc, the first publicly available Bangla--English code-mixed corpus for implicit meaning identification. The dataset contains 9,087 manually annotated sentences labeled for humor, sarcasm, offensiveness, and vulgarity. We construct the corpus through targeted social media collection, systematic filtering, and multi-annotator validation. We establish benchmark results using supervised transformer models and five frontier large language models (LLMs) evaluated in a zero-shot setting under structured prompting. Results show strong performance on humor detection, while sarcasm, offensiveness, and vulgarity remain substantially more challenging due to class imbalance and pragmatic complexity. The zero-shot LLMs provide competitive training-free baselines but achieve relatively low exact-match accuracy. Further analysis estimates that over 42\% of negative-sentiment instances in an external dataset exhibit sarcastic characteristics. MixSarc provides a foundational resource for culturally aware NLP and supports more reliable multi-label modeling in code-mixed environments.

\clearpage
\newgeometry{top=0.85in,left=1in,right=1in,footskip=0.75in}

\section*{Introduction}
In the rapidly evolving digital landscape of South Asia, code-mixing between Bangla and English has emerged as a prevalent mode of communication, particularly on social media platforms. Users seamlessly interweave Bangla (often in Romanized form, known as Banglish) with English words, phrases, slang, and culturally nuanced expressions. This fluid linguistic practice, while natural to millions of speakers, poses significant challenges to existing natural language processing (NLP) systems, which are predominantly trained on monolingual or high-resource language data.

Subjective tasks such as humor, sarcasm, offense, and vulgarity detection are especially difficult in code-mixed settings. These phenomena rely heavily on implicit cultural knowledge, contextual incongruity, and intentional ambiguity---elements that are frequently obscured by inconsistent transliterations, informal spellings, and intra-sentential language switching. Consequently, current sentiment analysis and toxicity detection models often misclassify sarcastic or humorous content as literal sentiment, or fail to identify veiled offense and vulgarity masked by irony.

Despite increasing work on code-mixed sentiment analysis for Indian languages, Bangla--English remains severely understudied, and virtually no resources exist for finer-grained subjective tasks like humor, sarcasm, offense, and vulgarity in this language pair (see Table~\ref{tab:related-datasets} for a comparison with existing datasets).

\begin{table}[!ht]
\centering
\caption{\textbf{Comparison of existing Bangla and Bangla--English code-mixed datasets relevant to sentiment, sarcasm, offense, and humor detection.}}
\label{tab:related-datasets}

\setlength{\tabcolsep}{4pt}
\renewcommand{\arraystretch}{1.1}

\begin{tabularx}{\linewidth}{@{}l X l r@{}}
\toprule
\textbf{Dataset} & \textbf{Language(s)} & \textbf{Task(s)} & \textbf{\#Samples} \\
\midrule
\textbf{BnSentMix}\cite{alam2024bnsentmix}
& Bangla-English Codemix
& Sentiment (4 labels)
& $\sim$20{,}000 \\

\textbf{BanglaSarc}\cite{apon2022banglasarc}
& Bangla
& Sarcasm detection
& 5{,}112 \\

\textbf{SentMix-3L}\cite{raihan2023mixed}
& Bangla-English-Hindi
& Sentiment analysis
& Not reported \\

\textbf{BanglishRev}\cite{shamael2024banglishrev}
& Bangla-English + Banglish
& Product review sentiment
& 1.74M \\

\textbf{BanTH}\cite{haider2025banth}
& Transliterated Bangla
& Multi-label hate speech
& 37{,}300 \\

\textbf{MixSarc (Ours)}
& Bangla--English Codemix
& Implicit meaning
& 9{,}087 \\
\bottomrule
\end{tabularx}
\end{table}

\begin{figure}[tbp]
\centering
\includegraphics[width=0.7\textwidth]{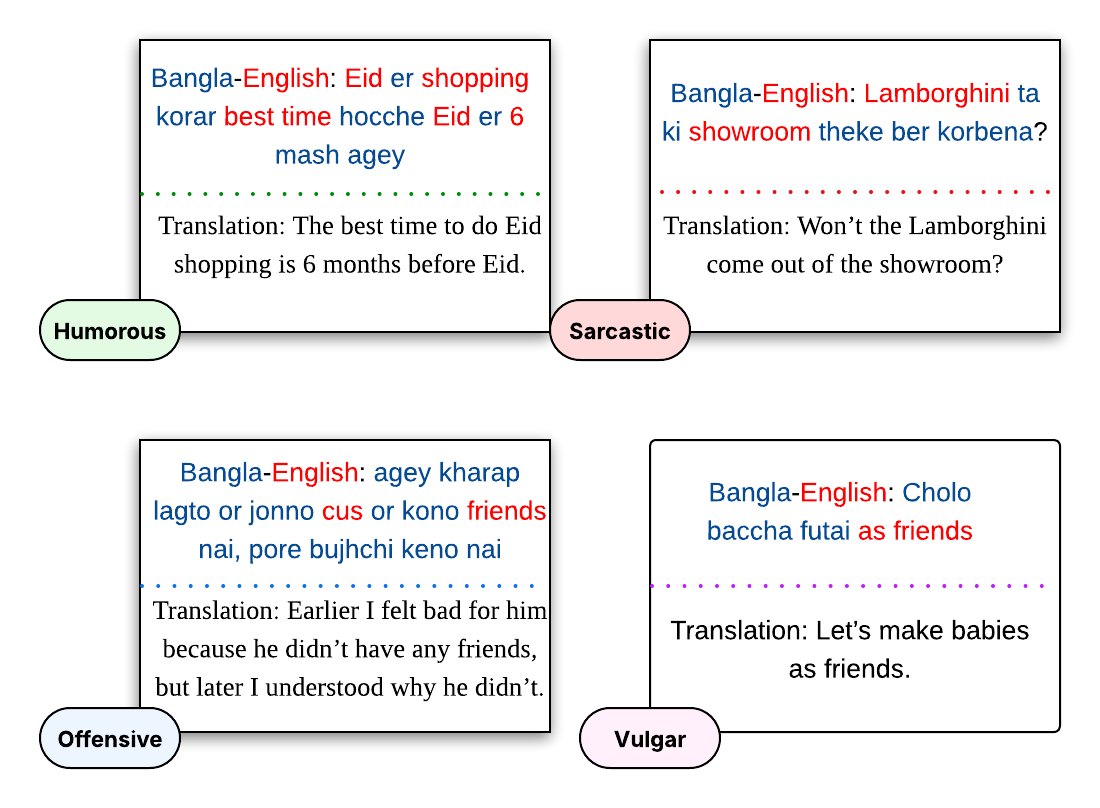}
\caption{\textbf{Examples of humorous, sarcastic, offensive, and vulgar utterances from the MixSarc dataset.}
Each example is presented in its original Bangla--English code-mixed form followed by an English translation. Red represents English words, blue represents Bengali words written in English alphabets.}
\label{fig:examples}
\end{figure}

\begin{figure}[t]
\centering
\includegraphics[width=0.7\textwidth]{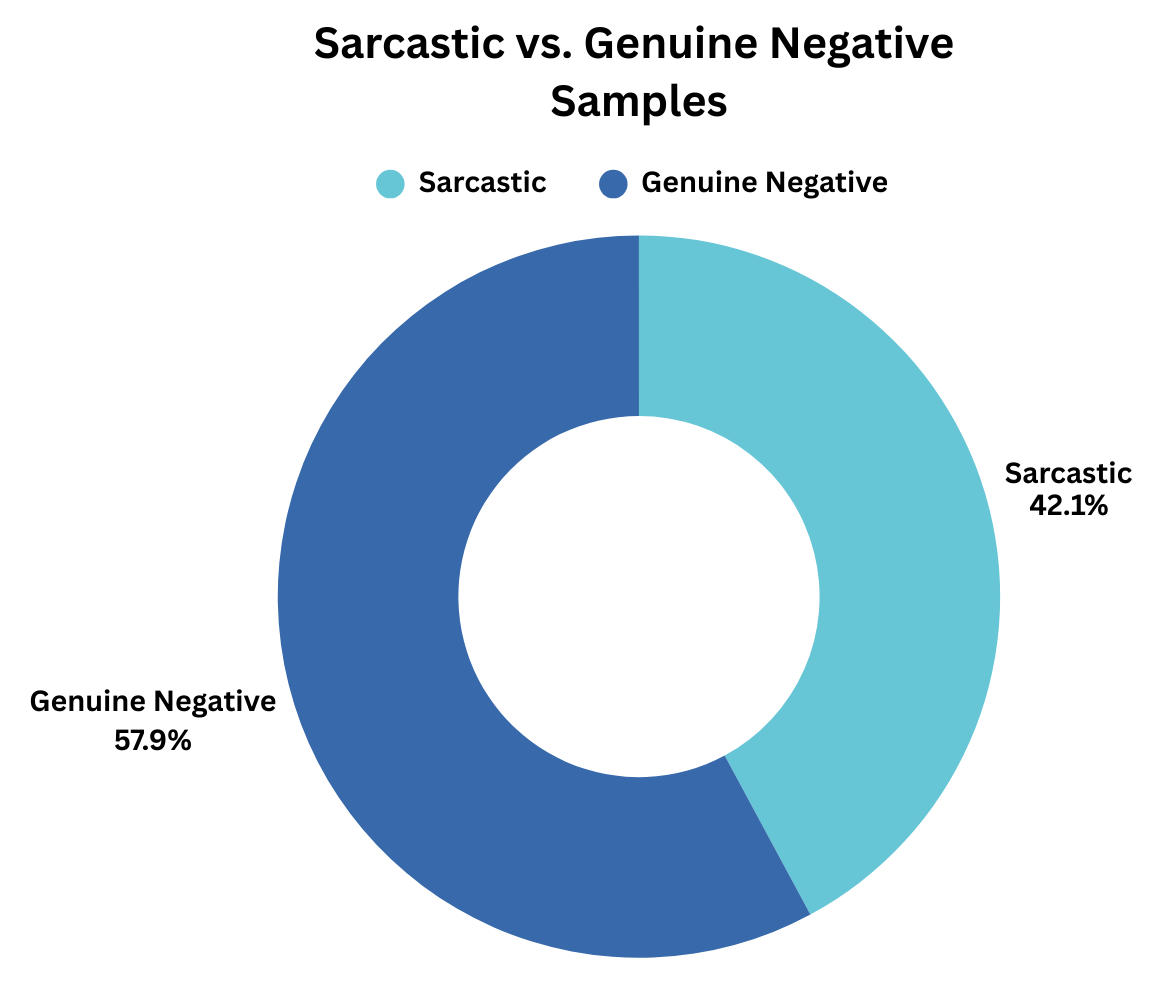}
\caption{Distribution of sarcastic vs. genuine negative samples within the negative sentiment class of BnSentMix.}
\label{fig:pieChart}
\end{figure}

To address this gap, we present \textbf{MixSarc}, the first publicly available, human-annotated Bangla--English code-mixed corpus specifically designed for humor, sarcasm, offense, and vulgarity detection. Collected from diverse online sources (YouTube comments, Facebook posts, and e-commerce reviews), MixSarc contains 9,087 samples annotated with four primary labels---Humorous, Sarcastic, Offensive, and Vulgar. Our contributions are threefold:

\begin{enumerate}
\item We introduce MixSarc, a rich, publicly available\footnote{The dataset is publicly available at \url{https://huggingface.co/datasets/ajwad-abrar/MixSarc}.} dataset comprising \textbf{9,087} Bangla--English code-mixed sentences that capture authentic linguistic and cultural nuances. Fig~\ref{fig:examples} shows representative examples from each category, highlighting how humor, sarcasm, offense, and vulgarity manifest in real-world code-mixed text.

\item We establish comprehensive benchmark results using classical machine learning classifiers, supervised transformer models (Banglish-BERT and Gemma-2B), and five frontier large language models evaluated in a zero-shot setting, reporting results across all four tasks.
    
\item Through in-depth analysis, we estimate that approximately 42.13\% of negative-sentiment instances in BnSentMix exhibit sarcastic characteristics (visualized in Fig.~\ref{fig:pieChart}), highlighting sarcasm as a frequent source of negative lexical cues in Bangla--English discourse. We further demonstrate that while transformer models perform well on humor detection, they struggle with sarcasm and completely fail on the minority classes of vulgarity and offense (F1 = 0.00) due to severe class imbalance.
\end{enumerate}

These findings highlight the necessity of imbalance-aware techniques (e.g., oversampling, class-weighted losses, or targeted augmentation) for robust detection of subjective toxicity in low-resource code-mixed settings. By releasing MixSarc along with detailed baselines and analyses, this work provides a solid foundation for future research on culturally aware NLP and safer content moderation for South Asian digital platforms.

\section*{Related Work}
Research on code-mixed language processing has expanded rapidly in recent years, driven by the widespread use of multilingual communication on South Asian social media. Bangla--English code-mixing, in particular, presents unique challenges for NLP due to inconsistent transliteration, non-standard grammar, and culturally embedded expressions that complicate language identification and downstream classification tasks.

\subsection*{Code-Mixed Language Processing}
Early work on Bangla--English code-mixed text focused on foundational tasks such as word-level language identification. Chanda et al.~\cite{chanda2016unraveling} introduced a predictor--corrector model combining rule-based heuristics and machine learning, supported by a Facebook chat corpus that remains a valuable early resource. More recently, Raihan et al.~\cite{raihan2023mixed} advanced code-mixed modeling through Mixed-DistilBERT, a lightweight transformer pretrained on both monolingual and synthetically generated code-mixed text. Their results demonstrated the effectiveness of task-specific pretraining even in low-resource, noisy environments.

\subsection*{Sentiment Analysis in Code-Mixed Text}
Bangla--English sentiment analysis has attracted increasing attention with the release of datasets such as SentMix-3L~\cite{raihan2023mixed} and BnSentMix~\cite{alam2024bnsentmix}. SentMix-3L introduced a tri-lingual Bangla--English--Hindi corpus combining natural and synthetic samples, highlighting the complexity of multilingual sentiment detection. BnSentMix provided a larger, more natural dataset with high-quality annotation and a “mixed’’ sentiment category to capture nuanced polarity expression. However, existing models trained on these datasets often misinterpret sarcastic or offensive content, revealing substantial performance degradation on subjective or context-heavy examples.

\begin{figure}[!ht]
  \centering
  \includegraphics[width=\textwidth]{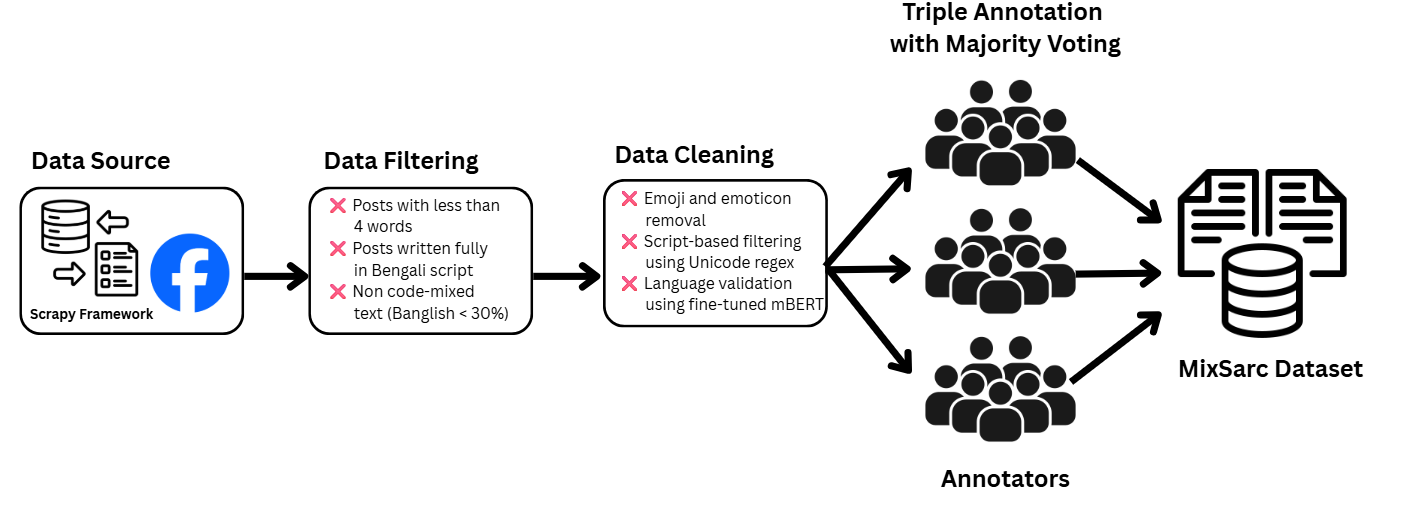}
  \caption{\textbf{Overview of the dataset preparation pipeline used in this work.}}
  \label{fig:dataset_pipeline}
\end{figure}


\subsection*{Sarcasm and Humor Detection}
Sarcasm detection has traditionally relied on detecting incongruity between surface sentiment and contextual meaning, as formalized by Riloff et al.~\cite{riloff2013sarcasm}. While substantial progress has been made in English--Hindi code-mixed sarcasm detection, including datasets by Swami et al.~\cite{swami2018corpus} and large-scale Hinglish corpora by Aggarwal et al.~\cite{aggarwal2020did}, Bangla--English sarcasm detection remains significantly underexplored. Bangla-only resources such as Ben-Sarc~\cite{lora2025ben} and BanglaSarc~\cite{apon2022banglasarc} provide early baselines but do not capture code-mixed linguistic behavior. Prior work on humor and offense detection using transformer-based systems, such as the DuluthNLP system~\cite{akrah2021duluthnlp}, further underscores the importance of contextual modeling and fine-tuning strategies.

\subsection*{Offensive and Vulgar Language Detection}
Work on detecting offensive language in Bangla--English code-mixed text is limited but growing. Mandal et al.~\cite{mandal2018preparing} introduced one of the earliest code mixed offensive corpora and demonstrated the effectiveness of traditional machine learning models using lexical features. Sazzed~\cite{sazzed2021abusive} contributed the BEOL dataset, distinguishing between offensive and vulgar expressions and showing the advantage of deep recurrent models over traditional techniques. More recent studies fine-tuned multilingual transformers such as mBERT for abusive content detection~\cite{sharif-etal-2021-nlp} and achieved improved performance, although challenges remain due to transliteration inconsistencies and subtle pragmatic cues.

\section*{Materials and methods}

\subsection*{MixSarc Dataset}
In this work, we introduce MixSarc, a Bangla--English code-mixed dataset designed for the joint detection of humor, sarcasm, offensiveness, and vulgarity in social media text. To the best of our knowledge, MixSarc is the first publicly available resource that simultaneously annotates these four pragmatic phenomena on transliterated Bangla--English content.

The dataset construction followed a multi-stage pipeline analogous to other recent code-mixed benchmarks (see Figure~\ref{fig:dataset_pipeline}): targeted data collection from public social media pages, systematic cleaning and filtering to retain linguistically robust code-mixed content, and a controlled annotation campaign with multiple trained annotators. Particular emphasis was placed on handling the variation in transliteration (Banglish), script mixing, and subtle context-dependent cues that are typical of informal online communication. In the following subsections, we present the data sourcing process, preprocessing and code mix filtering procedures, annotation protocol, label distribution, and the key challenges and limitations.

\subsection*{Data Sourcing}
To obtain naturally occurring humorous and sarcastic code-mixed text, we collected posts from publicly accessible Facebook pages that predominantly publish informal, culture-specific content in Bangla--English. Data collection was carried out using the \textit{Scrapy} framework\footnote{\url{https://github.com/scrapy/scrapy}}, which allowed us to systematically crawl and store page posts.

We selected four high-engagement Facebook pages as data sources and collected all available posts within a specified time window. \textbf{Table~\ref{tab:data_sources} summarizes the selected source pages and the number of posts collected from each page.}

\begin{table}[ht]
    \centering
    \begin{tabular}{l r}
        \hline
        \textbf{Source Page} & \textbf{Posts Collected} \\
        \hline
        \textit{Shelby Bhai} & 6,034 \\
        \textit{OneTwoThreeAmiOnekFree} & 8,315 \\
        \textit{Porte Bosh} & 7,386 \\
        \textit{Ammu Dake} & 9,394 \\
        \hline
    \end{tabular}
    \caption{Summary of data sources and number of posts collected.}
    \label{tab:data_sources}
\end{table}

In total, this step yielded \textbf{31,129} raw text items, which formed the initial pool for subsequent filtering and annotation.

\subsection*{Data Cleaning and Preprocessing}
Starting from the raw collection, we applied a series of cleaning and filtering operations to remove non-linguistic artifacts and retain posts that exhibit genuine Bangla--English code-mixing. The preprocessing pipeline consisted of emoji removal, script-based filtering, and a final code-mix validation stage based on a fine-tuned mBERT classifier.

\subsubsection*{Emoji and Non-Textual Cue Removal}
To eliminate non-textual cues and ensure that models trained on MixSarc rely solely on language-based signals, all emojis were removed using the \verb|replace_emoji| function from the Python  \verb|emoji|\footnote{\url{https://pypi.org/project/emoji/}} library. This ensures that humor, sarcasm, and offensiveness must be inferred from text rather than pictorial markers. The complete text cleaning procedure is summarized in Algorithm~\ref{alg:clean_text}.

\subsubsection*{Script-Based Filtering}

Because our focus is on transliterated Bangla and English, posts written entirely in Bangla script were discarded. We identified Bangla-only content by matching against Bangla Unicode ranges using regular expressions. Posts containing English script, Romanized Bangla (Banglish), or mixed scripts were preserved for downstream processing.

\begin{algorithm}[t]
\caption{Clean Text}
\label{alg:clean_text}
\begin{algorithmic}[1]
\Require $text \leftarrow$ Input text
\Ensure $text \leftarrow$ Preprocessed text

\State $text \leftarrow text.lower()$ \Comment{\{Convert to lowercase\}}
\State $text \leftarrow$ Remove all special characters except ``?'', ``,'', ``!'', and ``.''
\State $text \leftarrow$ Reduce consecutive sequences of punctuations to a single instance
\State $text \leftarrow$ Remove all non-ASCII characters
\State $text \leftarrow$ Remove extra white spaces
\State $text \leftarrow$ Capitalize the first letter after each period (.)
\State \Return $text$

\end{algorithmic}
\end{algorithm}

\subsection*{Code-Mixed Validation via mBERT}

To automatically verify that each post contains genuine Bangla--English code-mixing, we employed a token-level classifier derived from a fine-tuned Multilingual BERT (mBERT) model. The classifier assigns tokens to one of two categories:

\begin{itemize}
    \item \textbf{English}
    \item \textbf{Banglish} (Romanized Bangla)
\end{itemize}

Using these token-level predictions, we computed the proportion of Banglish tokens per post and applied a simple rule-based decision function. Let \verb|total_word_count| denote the number of tokens in the post and \verb|total_benglish_word_count| the number identified as Banglish. The full procedure is outlined in Algorithm~\ref{alg:code_mixed_detection}.

\begin{algorithm}[t]
\caption{Detect Code-mixed Bangla}
\label{alg:code_mixed_detection}
\begin{algorithmic}[1]

\Require $S \leftarrow$ List of sentences
\Require $model \leftarrow$ Pre-trained mBERT model
\Require $tokenizer \leftarrow$ Pre-trained mBERT tokenizer
\Ensure $pred \leftarrow$ Predicted class label (0 or 1)

\State $b\_count \leftarrow 0$
\State $w\_count \leftarrow 0$

\For{each $sent$ in $S$}
    \State $words \leftarrow split(sent)$
    \For{each $w$ in $words$}
        \State $w \leftarrow preprocess(w)$
        \If{$w$ is empty}
            \State \textbf{continue}
        \EndIf
        \State $w\_count \leftarrow w\_count + 1$
        \State $inputs \leftarrow tokenize(w)$
        \State $outputs \leftarrow model(inputs)$
        \State $pred\_class \leftarrow argmax(outputs)$
        \If{$pred\_class == 1$}
            \State $b\_count \leftarrow b\_count + 1$
        \EndIf
    \EndFor
\EndFor

\If{$w\_count < 4$}
    \State \Return 0
\EndIf

\State $b\_percent \leftarrow b\_count / w\_count$

\If{$b\_percent \geq 0.3$}
    \State \Return 1
\Else
    \State \Return 0
\EndIf

\end{algorithmic}
\end{algorithm}

Posts with fewer than four tokens were discarded to avoid extremely short and noisy content. Remaining posts were retained only if at least \textbf{30\%} of tokens were classified as Banglish, ensuring meaningful code-mixing. After this automatic filtering stage, a total of \textbf{9,087} high-quality code-mixed sentences were passed to the annotation phase.

\subsection*{Data Annotation}

\subsubsection*{Annotation Scheme}

Each sentence in MixSarc is annotated with four binary labels:

\begin{itemize}
    \item \textbf{Humorous} – indicates explicit or implicit comedic intent.
    \item \textbf{Sarcastic} – denotes irony, mockery, or contrast between literal and intended meaning.
    \item \textbf{Offensive} – marks potentially insulting, derogatory, or harmful expressions.
    \item \textbf{Vulgar} – captures coarse, profane, or obscene language usage.
\end{itemize}

The annotation is \textbf{multi-label}: a single sentence may simultaneously exhibit multiple phenomena (e.g., humorous and sarcastic, or offensive and vulgar). This design better reflects the overlapping nature of tone and intent in informal social media discourse.

Each sentence was independently annotated by \textbf{three} annotators. Final labels were determined through \textbf{majority voting} across annotators for each of the four binary dimensions. Detailed annotation guidelines, including definitions, labeling rules, and examples for all categories, are provided in the Appendix.

\subsubsection*{Annotators}

We recruited \textbf{18} annotators who are \textbf{native Bangla speakers}, fluent in Bangla, familiar with transliterated Bangla (Banglish), and comfortable with English. All annotators had formally studied Bangla as an academic subject up to at least the Higher Secondary level. Additionally, they were required to pass a qualification test, achieving at least \textbf{70\%} accuracy against a gold standard subset before participating in the main annotation process. 

\begin{table}[ht]
\centering
\begin{tabular}{l r}
\hline
\textbf{Attribute} & \textbf{Value} \\
\hline
Number of annotators & 18 \\
Age range & 19--59 years \\
Gender distribution & 11 male, 7 female \\
Countries represented & 6 \\
Institutions represented & 14 \\
Average sentences per annotator & $\sim$1500 \\
\hline
\end{tabular}
\caption{Annotator demographics and annotation workload.}
\label{tab:annotator_stats}
\end{table}

The annotator pool is demographically diverse as shown in Table~\ref{tab:annotator_stats}. On average, each annotator labeled approximately \textbf{1,500} sentences. Annotators were compensated upon completion of their assigned batches.

\subsubsection*{Label Statistics}

Table~\ref{tab:mixsarc_combined} reports the complete label distribution obtained after the full annotation process, along with the balanced subset used for model training. Each sentence in the corpus was annotated independently by three annotators, and the final label assignment was determined through majority voting. The resulting distribution, shown in Table~\ref{tab:mixsarc_combined} (column: Full Dataset), reflects the natural frequencies of humorous, sarcastic, offensive, vulgar, and neutral content across all 9,087 sentences, including all multi-label combinations. As shown, humorous (4,041) and sarcastic (1,475) expressions appear most frequently, while offensive (207) and vulgar (290) content occur less often. Multi-label overlaps such as \textit{humorous + sarcastic} (844) further highlight the intertwined nature of pragmatic cues in Bangla–English code-mixed discourse.

\subsection*{Dataset Statistics}

For downstream modeling, a balanced subset of the corpus was constructed to ensure equal representation across the primary single-label classes. Table~\ref{tab:mixsarc_combined} (column: Balanced Subset) shows the final curated counts used for training, validation, and testing. Each of the four primary labels—humorous, sarcastic, vulgar, and offensive—was capped at 750 instances, along with 750 neutral samples. Multi-label counts remain unchanged from the full dataset, preserving real-world co-occurrence patterns.

The final dataset contains 9,087 sentences and is split into training, validation, and test sets using a 70:15:15 ratio, resulting in 6,361 training samples, 1,363 validation samples, and 1,363 test samples.


\begin{table}[ht]
\centering
\setlength{\tabcolsep}{3pt} 
\caption{Full vs. balanced label distribution.}
\label{tab:mixsarc_combined}
\begin{tabular}{lrr}
\toprule
\textbf{Label} & \textbf{Full} & \textbf{Balanced} \\
\midrule
Humorous                     & 4041 & 750 \\
Sarcastic                    & 1475 & 750 \\
Offensive                    & 207  & 207 \\
Vulgar                       & 290  & 290 \\
None                         & 1953 & 750 \\
\midrule
Humor + Sarc.                & 844 & 844\\
Humor + Vul.                 & 84  & 84\\
Humor + Off.                 & 76  & 76\\
Off. + Sarc.                 & 38  & 38\\
Vul. + Off.                  & 34  & 34\\
Vul. + Sarc.                 & 27  & 27\\
Humor + Off. + Sarc.         & 9   & 9\\
Humor + Vul. + Sarc.         & 4   & 4\\
Humor + Vul. + Off.          & 3   & 3\\
Vul. + Off. + Sarc.          & 2   & 2\\
\bottomrule
\end{tabular}
\end{table}

\subsection*{Annotation Validation}

To assess annotation reliability, we computed Fleiss’ Kappa~\cite{fleiss1971measuring} to measure inter-annotator agreement among the three annotators. Unlike Cohen’s Kappa, which is limited to two raters, Fleiss’ Kappa extends the agreement measure to multiple annotators while correcting for chance agreement.

\subsubsection*{Inter-Annotator Agreement}

The overall agreement across the dataset yielded an average Fleiss’ Kappa coefficient of $\bar{\kappa} = 0.66$, corresponding to \textit{substantial agreement} under the conventional interpretation scale.

\subsubsection*{Factors Affecting Agreement}

The observed agreement level can be attributed to several inherent characteristics of the annotation task:

\begin{enumerate}
    \item \textbf{Implicit and Context-Sensitive Categories:}  
    The annotated labels—humor, sarcasm, vulgarity, and offensiveness—are inherently implicit and context-dependent. Unlike objective factual annotations, these categories require interpretation of pragmatic cues, tone, cultural references, and implied meanings, which increases annotation variability.

    \item \textbf{Annotator Diversity:}  
    Annotators were intentionally recruited from diverse age groups (19--59 years) and varied cultural and educational backgrounds. While this diversity improves ecological validity and dataset representativeness, it also introduces differences in perception regarding humor, sarcasm, and offensive content.

    \item \textbf{Intrinsic Subjectivity of Target Phenomena:}  
    The target phenomena are inherently subjective. Sarcasm often relies on irony and contextual incongruity; humor depends on personal and cultural preferences; and perceptions of offensiveness vary across individuals and social norms. Some variation in annotation is therefore expected even among trained annotators.
\end{enumerate}

Overall, the agreement level ($\bar{\kappa} = 0.66$) suggests reliable annotation quality while acknowledging the natural ambiguity associated with socially grounded and pragmatically nuanced language.

\section*{Methodology and Experimental Setup}

\subsection*{Modeling Approach}

We formulate the MixSarc classification task as a \textbf{multi-label text classification} problem, reflecting the fact that a single sentence may simultaneously express humor, sarcasm, offensiveness, and vulgarity. Each instance is represented by a four-dimensional binary vector:
\[
[y_{\text{hum}},\, y_{\text{sar}},\, y_{\text{off}},\, y_{\text{vul}}] \in \{0,1\}^4.
\]

To support this setup, the dataset was partitioned into 70\% training, 15\% validation, and 15\% test sets. Stratified sampling was applied using the humorous label to maintain distributional consistency. A modified PyTorch \texttt{Dataset} class was used to return tokenized inputs alongside multi-label vectors for each sample.

The classification models are based on transformer encoders, fine-tuned end-to-end on the MixSarc dataset. The use of pretrained language models allows the system to capture nuanced contextual and pragmatic cues that are characteristic of Bangla–English code-mixed discourse.

\subsection*{Evaluation Metrics}

We evaluate model performance using \textbf{classification accuracy}, \textbf{precision}, \textbf{recall}, and \textbf{F1-score}, following standard multi-label evaluation practice. All metrics are computed using \textbf{sample-based averaging}, which treats each sentence as an independent multi-label instance and averages performance across labels accordingly.

In addition, confusion matrices are generated separately for humor, sarcasm, offensiveness, and vulgarity, enabling a fine-grained analysis of misclassification patterns for each attribute.

\subsection*{Implementation Details}

All models were implemented using the HuggingFace Transformers library. Fine-tuning was performed using the following configuration:

\begin{itemize}
    \item \textbf{Loss Function:} Binary Cross-Entropy with Logits (\texttt{BCEWithLogitsLoss})
    \item \textbf{Optimizer:} AdamW
    \item \textbf{Learning Rate:} \(1.25 \times 10^{-6}\)
    \item \textbf{Scheduler:} Linear decay without warmup
    \item \textbf{Epochs:} 15
    \item \textbf{Batch Size:} 32
\end{itemize}

Training was conducted using GPU acceleration when available. During each iteration, the model performed a forward pass over input IDs and attention masks, computed independent binary losses for each label, aggregated them, and applied backpropagation followed by an optimization step. The training pipeline closely follows standard practices for multi-label fine-tuning of transformer architectures.

\subsection*{Zero-Shot Baseline with Large Language Models}

In addition to supervised fine-tuning, we evaluate the MixSarc task under a zero-shot learning setting using large language models (LLMs), with the goal of establishing strong, training-free baselines and analyzing the inherent capabilities and limitations of LLMs on Bangla--English code-mixed implicit meaning detection.

We benchmark five state-of-the-art frontier large language models (LLMs), comprising both proprietary and open-weight models, to evaluate their zero-shot performance on the MixSarc dataset.

Our selected five LLMs for assessment can be categorized into proprietary and open-source models:

\begin{enumerate}
    \item \textbf{Proprietary Models:}
    \begin{enumerate}
        \renewcommand{\labelenumii}{\roman{enumii}.}
        \item \textbf{Claude Opus 4.8:} Claude Opus 4.8 is a frontier model from Anthropic's Claude family, designed for complex reasoning, instruction following, and long-context language understanding. Although its parameter count is not publicly disclosed, Claude models are known for strong alignment, safety-focused training, and robust performance on nuanced language tasks, making the model suitable for zero-shot implicit meaning detection in code-mixed text~\cite{anthropic2026opus48}.

        \item \textbf{GPT-5.5:} GPT-5.5 is a proprietary model from OpenAI's GPT series and is designed for complex real-world tasks involving reasoning, information analysis, text generation, and tool-assisted workflows. While OpenAI does not disclose the exact architecture or parameter count, GPT-family models are widely used as strong general-purpose LLM baselines for language understanding and generation tasks~\cite{openai2026gpt55}.

        \item \textbf{Gemini 2.5 Flash:} Gemini 2.5 Flash is a lightweight and efficient model from Google's Gemini 2.5 family. It is optimized for fast inference while retaining strong instruction-following, multilingual understanding, and reasoning capabilities. Its efficiency makes it suitable for large-scale zero-shot benchmarking where cost and response time are important considerations~\cite{team2025gemini}.
    \end{enumerate}

    \item \textbf{Open-Source Models:}
    \begin{enumerate}
        \renewcommand{\labelenumii}{\roman{enumii}.}
        \item \textbf{Kimi K2 0711:} Kimi K2 is a Mixture-of-Experts (MoE) model developed by Moonshot AI, with 1 trillion total parameters and 32 billion activated parameters. It is trained for strong agentic, coding, reasoning, and instruction-following capabilities, and represents one of the strongest open-weight LLMs for evaluating zero-shot classification performance~\cite{kimi2025k2}.

        \item \textbf{LLaMA 4 Maverick:} LLaMA 4 Maverick is an open-weight model from Meta's LLaMA 4 family. It follows a Mixture-of-Experts architecture and is designed for advanced reasoning, multilingual understanding, and instruction-following tasks. Its open-weight availability and strong general-purpose capabilities make it a suitable representative of modern open-source LLMs in the zero-shot evaluation setting~\cite{dubey2025llama4}.
    \end{enumerate}
\end{enumerate}









\begin{tcolorbox}[
  colback=yellow!20,
  colframe=yellow!50!black,
  coltext=black,
  sharp corners,
  boxrule=0.5pt,
  left=5pt,
  right=5pt,
  top=5pt,
  bottom=5pt
]

\textbf{Zero-Shot Prompt for MixSarc Annotation}

Assign four binary labels (0 or 1) to the given sentence.

\textbf{Label Definitions:}
\begin{itemize}
    \item \textbf{Sarcastic}: Verbal irony with a mocking or insincere tone.
    \item \textbf{Humorous}: Intended to amuse or provoke laughter.
    \item \textbf{Offensive}: Contains insulting or identity-targeted language.
    \item \textbf{Vulgar}: Contains obscene or sexually explicit language.
\end{itemize}

Output only valid JSON with no explanations.

\textbf{Format:}

\verb|{"Humorous":0,"Sarcastic":0,"Offensive":0,"Vulgar":0}|

\textbf{Sentence:}

\verb|{text}|

\end{tcolorbox}

None of these models are fine-tuned on the MixSarc dataset. Instead, each instance is classified using a structured prompt designed to closely follow the original MixSarc annotation guidelines, ensuring consistency between human and model-based labeling.

The zero-shot task is formulated as a \textbf{multi-label binary classification} problem consistent with the supervised setup. Given a single sentence, the model is instructed to assign four binary labels corresponding to \textit{Humorous}, \textit{Sarcastic}, \textit{Offensive}, and \textit{Vulgar}. The prompt defines each category and enforces strict output constraints. Models are required to return predictions in valid JSON format with binary values \(0\) and \(1\) for each label, ensuring deterministic parsing and compatibility with automated evaluation. 

To reduce randomness and improve consistency, the decoding temperature is set to zero for all models. Inference is performed on the held-out test split using the OpenRouter API. For each sentence, the generated prediction is parsed into a four-dimensional binary vector corresponding to the four target labels, directly matching the ground-truth annotation format.

Evaluation is conducted on the full test split using \textbf{exact match accuracy}, \textbf{precision}, \textbf{recall}, and \textbf{F1-score}. In addition to reporting overall performance, label-wise metrics are computed for each of the four categories. Exact match accuracy measures the proportion of samples for which all four labels are predicted correctly, providing a strict assessment of holistic multi-label performance. Precision, recall, and F1-score are reported using macro-averaged aggregation across labels, ensuring that each category contributes equally to the final evaluation regardless of class distribution.

\begin{table}[!t]
\centering
\caption{
Performance of supervised models and zero-shot LLM baselines on the MixSarc dataset. Highlighted cells indicate the best score within each category.
}
\label{tab:model_comparison}
\begin{tabular}{llcccc}
\toprule
\textbf{Approach} & \textbf{Model} & \textbf{Acc.} & \textbf{Prec.} & \textbf{Rec.} & \textbf{F1} \\
\midrule
\multicolumn{6}{c}{\textbf{Humor}} \\
\midrule
Supervised & Banglish-BERT     & \textbf{0.6232} & 0.6230 & 0.8197 & \textbf{0.7080} \\
Supervised & Gemma-2B          & 0.6012 & 0.6007 & \textbf{0.8474} & 0.7031 \\
Zero-shot  & Claude Opus 4.8   & 0.5968 & 0.6235 & 0.6354 & 0.6294 \\
Zero-shot  & GPT-5.5           & 0.6078 & 0.6420 & 0.6150 & 0.6282 \\
Zero-shot  & Kimi K2 0711           & 0.5587 & 0.6177 & 0.4748 & 0.5369 \\
Zero-shot  & Gemini 2.5 Flash  & 0.5689 & \textbf{0.6861} & 0.3687 & 0.4796 \\
Zero-shot  & LLaMA 4 Maverick  & 0.5374 & 0.5644 & 0.6204 & 0.5911 \\
\midrule
\multicolumn{6}{c}{\textbf{Sarcasm}} \\
\midrule
Supervised & Banglish-BERT     & 0.6569 & 0.3689 & 0.4222 & \textbf{0.3938} \\
Supervised & Gemma-2B          & \textbf{0.7287} & \textbf{0.4539} & 0.0944 & 0.1553 \\
Zero-shot  & Claude Opus 4.8   & 0.6004 & 0.3124 & 0.4278 & 0.3611 \\
Zero-shot  & GPT-5.5           & 0.6672 & 0.3025 & 0.2000 & 0.2408 \\
Zero-shot  & Kimi K2 0711          & 0.5946 & 0.2916 & 0.3750 & 0.3281 \\
Zero-shot  & Gemini 2.5 Flash  & 0.6818 & 0.2874 & 0.1389 & 0.1873 \\
Zero-shot  & LLaMA 4 Maverick  & 0.4714 & 0.2807 & \textbf{0.6417} & 0.3905 \\
\midrule
\multicolumn{6}{c}{\textbf{Offensive}} \\
\midrule
Supervised & Banglish-BERT     & 0.9508 & 0.1250 & 0.0364 & 0.0563 \\
Supervised & Gemma-2B          & \textbf{0.9589} & 0.0000 & 0.0000 & 0.0000 \\
Zero-shot  & Claude Opus 4.8   & 0.8570 & 0.2140 & \textbf{0.6389} & \textbf{0.3206} \\
Zero-shot  & GPT-5.5           & 0.8710 & 0.2143 & 0.5417 & 0.3071 \\
Zero-shot  & Kimi K2 0711          & 0.9142 & \textbf{0.2353} & 0.2778 & 0.2548 \\
Zero-shot  & Gemini 2.5 Flash  & 0.9076 & 0.1786 & 0.2083 & 0.1923 \\
Zero-shot  & LLaMA 4 Maverick  & 0.9164 & 0.2083 & 0.2083 & 0.2083 \\
\midrule
\multicolumn{6}{c}{\textbf{Vulgar}} \\
\midrule
Supervised & Banglish-BERT     & \textbf{0.9509} & \textbf{0.5000} & 0.1194 & 0.1928 \\
Supervised & Gemma-2B          & 0.9509 & 0.5000 & 0.0299 & 0.0563 \\
Zero-shot  & Claude Opus 4.8   & 0.9326 & 0.3373 & \textbf{0.4308} & \textbf{0.3784} \\
Zero-shot  & GPT-5.5           & 0.9172 & 0.2500 & 0.3692 & 0.2981 \\
Zero-shot  & Kimi K2 0711         & 0.9311 & 0.3014 & 0.3385 & 0.3188 \\
Zero-shot  & Gemini 2.5 Flash  & 0.9238 & 0.2809 & 0.3846 & 0.3247 \\
Zero-shot  & LLaMA 4 Maverick  & 0.9362 & 0.2500 & 0.1692 & 0.2018 \\
\bottomrule
\end{tabular}
\end{table}

\section*{Result Analysis}

\subsection*{Benchmarking Transformer Models}

We evaluate the performance of two transformer-based architectures---Banglish-BERT and Gemma-2B---on the MixSarc dataset across four binary classification tasks: Humor, Sarcasm, Vulgarity, and Offensiveness. The comparison uses Accuracy, Precision, Recall, and F1-score to capture overall and class-sensitive performance. The detailed results are presented in Table~\ref{tab:model_comparison}.

\subsubsection*{Banglish-BERT Performance}

Banglish-BERT\footnote{\url{https://huggingface.co/aplycaebous/tb-BanglaBERT-fpt}} achieves its strongest results on \textbf{Humor}, with an F1-score of 0.708 and high recall (0.8197). This suggests the model is sensitive to humorous cues, though moderate precision implies false positives remain present.

Sarcasm detection demonstrates significantly lower performance (F1 = 0.3938), highlighting the inherent difficulty of interpreting ironic and context-dependent expressions in code-mixed text.

For Vulgarity and Offensiveness, the model shows high accuracy but very low recall and F1-scores, largely due to strong label imbalance. Although the model predicts the majority class reliably, minority-class examples are frequently missed.

\subsubsection*{Gemma-2B Performance}
Gemma-2B\footnote{\url{https://huggingface.co/google/gemma-2b}}, fine-tuned using QLoRA (4-bit quantization with Low-Rank Adaptation adapters), shows comparable performance to Banglish-BERT on Humor, achieving an F1-score of 0.7031 with high recall (0.8474). 

In Sarcasm detection, Gemma-2B attains higher accuracy and precision but extremely low recall (0.0944), indicating a conservative prediction strategy that misses most sarcastic instances. 

Performance on Vulgarity and Offensiveness mirrors Banglish-BERT: extremely low recall and F1-scores despite high overall accuracy. Particularly, the Offense task yields an F1-score of 0, meaning Gemma-2B fails to correctly identify any offensive examples.

\subsection*{Comparative Insights}
Both models handle humor effectively, driven by strong recall. Sarcasm remains the most challenging task, with neither model achieving satisfactory F1-scores. Banglish-BERT displays a more balanced precision–recall pattern, while Gemma-2B prioritizes precision at the expense of recall.

On Vulgar and Offensive content, the nearly zero recall values indicate that the rarity and contextual subtlety of these categories severely hinder transformer-based classification. These results emphasize the need for:
\begin{itemize}
    \item better class balancing,
    \item domain-specific augmentation strategies,
    \item more context-aware architectures for minority-class detection.
\end{itemize}

\subsection*{Zero-Shot Evaluation}
Table~\ref{tab:model_comparison} reports the performance of five zero-shot LLM baselines: Claude Opus 4.8, GPT-5.5, Kimi K2, Gemini 2.5 Flash, and LLaMA 4 Maverick. All models are evaluated through OpenRouter-based API\footnote{\url{https://openrouter.ai/}} inference on the held-out test split. Since the task is multi-label, each model predicts four binary labels for every sentence: \textit{Humorous}, \textit{Sarcastic}, \textit{Offensive}, and \textit{Vulgar}. We report label-wise accuracy, precision, recall, and F1-score for each category.

Overall, the zero-shot models achieve their strongest performance on the \textit{Humor} label. Claude Opus 4.8 and GPT-5.5 obtain the highest humor F1-scores, with values of 0.6294 and 0.6282, respectively. LLaMA 4 Maverick also performs competitively on humor detection, achieving an F1-score of 0.5911. This suggests that current instruction-following LLMs can identify humorous cues in Bangla--English code-mixed text reasonably well without task-specific fine-tuning.

For \textit{Sarcasm}, performance is lower across all models, reflecting the difficulty of detecting implicit, ironic, and culturally grounded meanings. LLaMA 4 Maverick achieves the highest sarcasm F1-score of 0.3905, mainly due to its high recall of 0.6417. Claude Opus 4.8 follows with an F1-score of 0.3611. In contrast, GPT-5.5 and Gemini 2.5 Flash show more conservative behavior, producing lower recall and therefore lower F1-scores.

The \textit{Offensive} and \textit{Vulgar} categories remain challenging due to class imbalance and the subtle nature of toxicity in code-mixed text. Claude Opus 4.8 performs best on both categories, achieving an F1-score of 0.3206 for offensiveness and 0.3784 for vulgarity. However, the overall scores remain modest, indicating that even advanced LLMs struggle to reliably identify minority-class pragmatic labels in zero-shot settings.

These results highlight two key observations. First, zero-shot LLMs provide useful training-free baselines for MixSarc, especially for humor detection. Second, performance varies substantially across labels, with sarcasm, offensiveness, and vulgarity remaining difficult without supervised adaptation. This reinforces the importance of task-specific fine-tuning, imbalance-aware training, and culturally grounded modeling for reliable multi-label prediction in Bangla--English code-mixed NLP.

\subsection*{Improving Sentiment Analysis Through Sarcasm Detection}

\subsubsection*{Revisiting Sentiment Labels in BnSentMix}

To evaluate the broader applicability of MixSarc-based models, we applied a MixSarc-finetuned BERT model to the BnSentMix dataset. Since sarcasm often manifests with negative lexical cues, traditional sentiment systems frequently misclassify sarcastic sentences as genuinely negative.

\subsubsection*{Findings}

Our analysis of the negative class in BnSentMix reveals that a substantial portion of samples are likely sarcastic. The MixSarc-tuned classifier estimates a sarcasm probability of 0.4213 for negative-labeled sentences. Out of 6,172 negative samples, approximately 2,600 exhibit sarcastic characteristics.

\subsubsection*{Implications}

These results highlight sarcasm as a critical confounding factor in sentiment analysis:
\begin{itemize}
    \item \textbf{Reliability Improvement:} Correcting sarcastic misclassifications improves sentiment accuracy for downstream tasks such as opinion mining.
    \item \textbf{Dataset Insight:} The presence of sarcasm in over 40\% of negative samples exposes a significant limitation in current annotation practices.
    \item \textbf{Practical Value:} In real-world social media and customer feedback systems, distinguishing sarcastic negativity from genuine negativity is essential to avoid misinterpretation of user intent.
\end{itemize}

Overall, this case study reinforces the practical significance of MixSarc’s multi-label framework for improving sentiment reliability in multilingual, code-mixed domains.

\section*{Conclusion}
This work introduced MixSarc, the first publicly available Bangla–English code-mixed dataset for implicit meaning identification. The dataset jointly covers humor, sarcasm, offensiveness, and vulgarity. It addresses an important gap in resources for studying pragmatic phenomena in Bangla–English social media. The annotation process followed structured guidelines and achieved substantial inter-annotator agreement, supporting the reliability of the dataset.

We also established benchmark results using supervised transformer models and frontier large language models in a zero-shot setting. Humor was the easiest category to detect, while sarcasm, offensiveness, and vulgarity remained challenging. These results highlight the impact of implicit meaning, cultural context, and class imbalance on model performance. They also demonstrate the need for more robust methods for code-mixed language understanding.

By releasing MixSarc together with baseline results, we provide a valuable resource for future research on Bangla–English code-mixed NLP and implicit meaning identification. We hope this benchmark encourages the development of more reliable, culturally aware, and pragmatics-aware language models for low-resource multilingual settings.

\bibliography{plos_bibtex_sample}

\clearpage
\appendix

\section*{Annotation Guidelines for MixSarc}
\label{sec:appendix}

This appendix presents the annotation guidelines used to construct the \textbf{MixSarc} corpus. 
Each sentence in the dataset is annotated along four pragmatic dimensions:

\begin{enumerate}
    \item \textbf{Sarcasm} (Yes / No)
    \item \textbf{Humor} (Yes / No)
    \item \textbf{Offense} (Yes / No)
    \item \textbf{Vulgarity} (Yes / No)
\end{enumerate}

Every sentence must receive a label (\texttt{Yes} or \texttt{No}) for \emph{each} of the four criteria. 
Annotations are \emph{multi-label}: a sentence may be humorous and sarcastic at the same time, 
or offensive and vulgar, etc. Annotators are instructed not to skip any sentence.

\subsection*{Sarcasm}

\subsubsection*{Definition.}
Sarcasm is a form of verbal irony where the literal meaning of an utterance is 
opposite to the intended meaning, often with a mocking or ridiculing tone.

\subsubsection*{Labeling Rules.}
Mark \textbf{Sarcasm = Yes} if:
\begin{itemize}
    \item The sentence appears positive on the surface but implies a negative meaning.
    \item There is exaggerated, insincere, or over-the-top praise.
    \item The tone is clearly mocking or ridiculing someone or something.
\end{itemize}

Mark \textbf{Sarcasm = No} if:
\begin{itemize}
    \item The sentence is direct, literal, or straightforward.
    \item The sentence may be ironic but does not contain mockery or ridicule.
\end{itemize}

\subsubsection*{Examples.}
\begin{itemize}
    \item \emph{``Wow! Ki boro genius re tor matha ektu beshi kaj kore!''} \\
    (Over-the-top praise implying the opposite) $\Rightarrow$ \textbf{Sarcasm = Yes}
    \item \emph{``Ami class e ashlam on-time, teacher impressed hoye amake bari pathai dilo!''} \\
    (Apparently positive, but clearly ironic) $\Rightarrow$ \textbf{Sarcasm = Yes}
    \item \emph{``Ami kalke porashona korinai.''} \\
    (Literal statement, no mockery) $\Rightarrow$ \textbf{Sarcasm = No}
\end{itemize}

\subsection*{Humor}

\subsubsection*{Definition.}
Humor refers to language that is intended to be funny, amusing, or entertaining, 
for example through jokes, puns, or playful exaggeration.

\subsubsection*{Labeling Rules.}
Mark \textbf{Humor = Yes} if:
\begin{itemize}
    \item The sentence contains jokes, puns, or playful exaggerations.
    \item The primary intention is to provoke laughter or amusement.
\end{itemize}

Mark \textbf{Humor = No} if:
\begin{itemize}
    \item The content is serious, angry, neutral, or purely informative.
    \item The content is sarcastic but \emph{not} clearly meant to be funny.
\end{itemize}

\subsubsection*{Examples.}
\begin{itemize}
    \item \emph{``Friend-zoned hoar jonno ami ekta premium badge paowa uchit.''} \\
    (Self-deprecating joke) $\Rightarrow$ \textbf{Humor = Yes}
    \item \emph{``Tumi ashbe na jani, tai ami aram kore bose asi.''} \\
    (Neutral, not clearly humorous) $\Rightarrow$ \textbf{Humor = No}
\end{itemize}

\subsection*{Offense}

\subsubsection*{Definition.}
Offense covers language that insults, demeans, or attacks a person or group. 
This includes name-calling, slurs, or hate speech directed at individuals or communities.

\subsubsection*{Labeling Rules.}
Mark \textbf{Offense = Yes} if:
\begin{itemize}
    \item The sentence contains aggressive insults or explicit verbal attacks.
    \item It targets someone’s identity (e.g., race, gender, religion, appearance, or other protected traits).
\end{itemize}

Mark \textbf{Offense = No} if:
\begin{itemize}
    \item The sentence is critical, emotional, or negative but does not personally attack a person or group.
    \item The sentence expresses anger or frustration without direct insult or demeaning language.
\end{itemize}

\subsubsection*{Examples.}
\begin{itemize}
    \item \emph{``Tui eto kala keno?''} \\
    (Insult targeting physical appearance/skin color) $\Rightarrow$ \textbf{Offense = Yes}
    \item \emph{``Eder shobai chagol, kono akal nai!''} \\
    (Demeaning a group with animal comparison) $\Rightarrow$ \textbf{Offense = Yes}
    \item \emph{``Ami or upor ragi, kintu kichu boli nai.''} \\
    (Expressing anger without insult) $\Rightarrow$ \textbf{Offense = No}
\end{itemize}

\subsection*{Vulgarity}

\subsubsection*{Definition.}
Vulgarity refers to obscene, sexually explicit, or profane language that is 
inappropriate in formal or public settings. It focuses on \emph{lexical} vulgarity, 
not just rude tone.

\subsubsection*{Labeling Rules.}
Mark \textbf{Vulgarity = Yes} if:
\begin{itemize}
    \item The sentence includes crude or explicit references to sex or body parts.
    \item It uses slang, swear words, or profane expressions with strong inappropriate connotations.
\end{itemize}

Mark \textbf{Vulgarity = No} if:
\begin{itemize}
    \item The language is rude or harsh but not sexually explicit or profane.
    \item There are insults or offensive remarks without vulgar or obscene terms.
\end{itemize}

\subsubsection*{Examples.}
\begin{itemize}
    \item \emph{``Tui ekdom bokachoda!''} \\
    (Contains vulgar slur) $\Rightarrow$ \textbf{Vulgarity = Yes}
    \item \emph{``O pura sex joke-er factory.''} \\
    (Explicit sexual reference) $\Rightarrow$ \textbf{Vulgarity = Yes}
    \item \emph{``Tor matha thik ase?''} \\
    (Rude but not obscene) $\Rightarrow$ \textbf{Vulgarity = No}
\end{itemize}

\subsection*{Final Notes}

For each sentence, annotators must:
\begin{itemize}
    \item Assign a \texttt{Yes} or \texttt{No} label for \emph{all four} dimensions:
    Sarcasm, Humor, Offense, and Vulgarity.
    \item Allow multiple labels to be \texttt{Yes} simultaneously (e.g., a sentence can be both 
    sarcastic and humorous, or both offensive and vulgar).
    \item Avoid leaving any sentence unlabeled.
\end{itemize}

If an annotator is uncertain about a particular case, they are encouraged to flag the instance 
for further review or discuss it with a supervisor. Consistent application of these guidelines 
is essential to ensure the reliability and usefulness of the MixSarc dataset for future research.

\end{document}